\documentclass{article}

\usepackage[preprint]{neurips_2025}

\usepackage[utf8]{inputenc} 
\usepackage[T1]{fontenc}    
\usepackage{hyperref}       
\usepackage{url}            
\usepackage{booktabs}       
\usepackage{amsfonts}       
\usepackage{nicefrac}       
\usepackage{microtype}      
\usepackage{xcolor}         
\usepackage{enumitem}
\usepackage{amsmath}
\usepackage{graphicx}
\usepackage{float}
\usepackage{multirow}
\usepackage{tabularx}
\usepackage{ragged2e}
\newcolumntype{C}{>{\Centering}X} 

\title{Are classical deep neural networks weakly adversarially robust?}

\newcommand{\mailto}[1]{%
  \begingroup
  \hypersetup{pdfborder={0 0 0}, colorlinks=false}%
  \href{mailto:#1}{\texttt{#1}}%
  \endgroup
}
\renewcommand{\thefootnote}{\fnsymbol{footnote}}
\author{%
  Nuolin~Sun\footnotemark[1],  Linyuan~Wang\footnotemark[1],  Dongyang~Li,  Bin~Yan,   Lei~Li\footnotemark[2] \\
  \texttt{\mailto{sunnuolinsnl@163.com}},  \texttt{\mailto{wanglinyuanwly@163.com}} \\
  \footnotetext[1]{Equal contribution}  
  \footnotetext[2]{Corresponding author} 
  Information Engineering University, Zhengzhou, China
}

\begin{document}

\maketitle
\begingroup
\renewcommand{\thefootnote}{\fnsymbol{footnote}} 
\footnotetext[1]{Equal contribution}
\footnotetext[2]{Corresponding author}
\endgroup

\begin{abstract}

Adversarial attacks have received increasing attention and it has been widely recognized that classical DNNs have weak adversarial robustness. The most commonly used adversarial defense method, adversarial training, improves the adversarial accuracy of DNNs by generating adversarial examples and retraining the model. However, adversarial training requires a significant computational overhead. In this paper, inspired by existing studies focusing on the clustering properties of DNN output features at each layer and the Progressive Feedforward Collapse phenomenon, we propose a method for adversarial example detection and image recognition that uses layer-wise features to construct feature paths and computes the correlation between the examples feature paths and the class-centered feature paths. Experimental results show that the recognition method achieves 82.77\% clean accuracy and 44.17\% adversarial accuracy on the ResNet-20 with PFC. Compared to the adversarial training method with 77.64\% clean accuracy and 52.94\% adversarial accuracy, our method exhibits a trade-off without relying on computationally expensive defense strategies. Furthermore, on the standard ResNet-18, our method maintains this advantage with respective metrics of 80.01\% and 46.1\%. This result reveals inherent adversarial robustness in DNNs, challenging the conventional understanding of the weak adversarial robustness in DNNs.

\end{abstract}

\section{Introduction}
\label{Introduction}
Since the introduction of adversarial examples \cite{1,2}, deep neural networks have faced the security threat of adversarial attacks. The most widely used defense strategy currently is adversarial training \cite{3,4}. Although adversarial attack and defense methods have engaged in continuous advancement in the arms race, defensive strategies persistently remain in a passive position overall \cite{5,6,7,8}. Further theoretical studies have shown that adversarial examples mainly utilize the non-robust features extracted by DNNs, making the weak adversarial robustness in DNNs more consistently accepted in the field \cite{9}.

Karkar models DNNs as discrete dynamical systems for adversarial example detection \cite{10}. The approach clusters layer-wise output features,  utilizing discriminative cluster-based features to separate clean and adversarial examples. The layer-wise clustering of output features reminds us of Neural Collapse (NC) studies \cite{11,12,13}, particularly recent advances in Progressive Feedforward Collapse (PFC) \cite{11}. PFC demonstrates that standard DNNs (e.g., ResNet) can exhibit layer-wise neural collapse after extended training epochs, where the output features of the final few layers consistently demonstrate excellent clustering properties. Specifically, normal examples features in these layers converge tightly around their respective class centroids. PFC includes: (1) layer-wise variability collapse (PFC1), (2) distance between class means and ETF (PFC2), and (3) layer-wise nearest class center accuracy (PFC3).

This inspires us to obtain a straightforward priori assumption for discriminating clean examples from adversarial examples: We assume that the layer-wise clustered features from DNNs form a unique feature path, which we define as a layer-wise feature path. And the class center can also correspond to one such layer-wise feature path. Clean examples show high correlation with their true class path, especially in the final few layers with PFC. In contrast, adversarial examples transition from their original class to incorrect class, leading to lower correlation with any class-specific path. Based on this hypothesis, we propose an adversarial example detection method using layer-wise feature paths correlations, achieving over 80\% adversarial detection accuracy on CIFAR-10. Moreover, we revisit the problem of adversarial example recognition. If adversarial examples can already be accurately detected with high probability, the final few layers of their layer-wise feature paths will be close to the wrong class. We choose the relatively forward layers to examine the clustering features and then give the recognition method, which achieves 44.17\% adversarial accuracy . To further demonstrate that our method provides novel insights into the intrinsic properties of DNN architectures, we additionally evaluated the standard ResNet-18 without any specialized training or modifications. Results show that our method achieves over 80\% clean accuracy and over 45\% adversarial accuracy on CIFAR-10, significantly outperforming the previous robustness of ResNet-18, which dropped to below 5\% under adversarial attacks.

The main contributions of this work are summarized as followed:

\begin{itemize}[leftmargin=*]
\item[$\bullet$] We propose a method for adversarial example detection and image recognition based on layer-wise feature paths of DNNs, which exploits the potential adversarial robustness of DNNs. On both network with PFC and standard ResNet-18, we achieve over 40\% adversarial accuracy on CIFAR-10, challenging the conventional understanding of weak adversarial robustness in DNNs.

\item[$\bullet$] Our method does not utilize any priori information for adversarial example attacks, nor does it use high computational overhead defense strategies similar to adversarial training. Compared to conventional adversarial training, which achieves 78.48\% clean accuracy and 50.74\% adversarial accuracy, our method attains a trade-off with 80.01\% clean accuracy and 46.1\% adversarial accuracy without relying on computationally expensive defense strategies.
\end{itemize}

\section{Method}
\label{section:2}
In this section, we propose a method for adversarial example detection and image recognition based on layer-wise feature paths. By capturing examples through each residual block output feature of the network to construct layer-wise feature paths, we evaluate the cosine similarity between the layer-wise feature paths of test examples and class-centered layer-wise feature paths for all classes. For the detection task, clear separability of layer-wise feature paths and similarity thresholding are required to identify adversarial examples. While recognition tasks prioritize identifying the feature path point most similar to the correct class as the key objective, we design an intermediate-layer weighted voting decision mechanism to determine the most robust intermediate layers.

\subsection{Layer-wise feature paths}
\label{section:21}
\begin{figure}[htbp]
\centering
\includegraphics[width=0.8\textwidth]{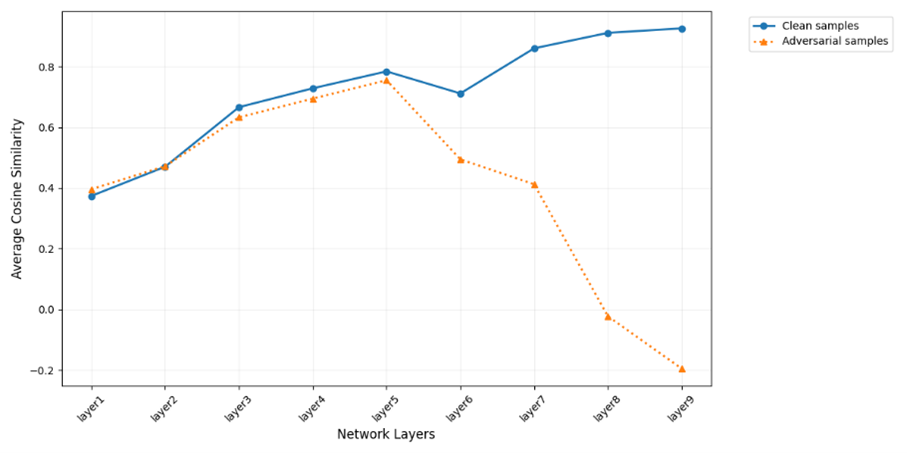}
\caption{Layer-wise feature paths comparison between clean and adversarial examples based on ResNet-20}
\label{fig:311_1}
\end{figure}

We define layer-wise feature paths to characterize the differences between adversarial examples and clean examples. In \cite{10}, the authors propose constructing a transmission trajectory using the features of individual examples passing through each layer of the DNN and perform adversarial example detection based on it. We analyze and find that the transmission trajectory exhibits good properties if the features obtained after the forward propagation of each example through each layer of the DNN remain stable. This leads us to associate PFC, a new development related to Neural Collapse \cite{11}. Neural Collapse can not only appear in the last layer, but can also be extended to the previous layers. This means that DNN with PFC will exhibit NC properties in the last layers of the network, i.e., the output features of the examples in the final few layers will all be stably clustered around the class center. We thus propose a method to portray differences between adversarial examples and clean examples: We define that a single example clustered through the outputs of each layer of DNN can obtain a feature path formed by layer-wise features, and the centers of each class layer-wise can also form such a feature path. For clean and adversarial examples, in Fig. \ref{fig:311_1}, it is not difficult to imagine that the features in each layer of the clean examples maintain a high correlation with the class center, whereas adversarial examples will show a higher correlation with the wrong class and a lower correlation with the correct class in the last layers where PFC appears. Therefore, we give a new method for adversarial example detection and image recognition under the definition of layer-wise feature paths. To illustrate the generalizability of this study, we also further extend experiments to standard ResNet-18 by constructing the same layer-wise feature paths.

Considering that existing attack methods mainly focus on directional perturbation and cosine similarity is sufficient to capture differences in layer-wise feature paths, we use the cosine similarity of example features to characterize the distance of layer-wise feature paths between examples \cite{10}. Furthermore, we define the number of layers for feature path statistics based on the residual blocks.

Based on PFC metrics, class centers simplify high-dimensional feature distributions into single representative vectors. This low-dimensional statistical compression inherently reflects geometric separability and orthogonality in feature spaces. Therefore, during model training, we construct layer-wise feature paths for all examples within the same class by utilizing class centers.

We first compute the feature vectors of each training example across each DNN layer. Next, we calculate the average features per class and normalize them. The class centers $\mu _{c}^{l}$ form layer-wise feature vectors $\mu$, constructing a feature path as
\begin{equation} \label{eq:1}
\mu _{c}^{l}=\text{Norm}(\frac{1}{{{n}_{c}}}\sum\limits_{i=1}^{{{n}_{c}}}{h_{c,i}^{l}})
\end{equation}
where $h_{c,i}^{l}\in {{\mathbb{R}}^{d}}$ denote the feature vector of the i-th training example in class $c$ at layer $l$. ${{{n}_{c}}}$ represents the number of training examples in class $c$. The L2 normalization operation $\operatorname { N o r m } ( \mathbf { x } ) = \frac { \mathbf { x } } { \| \mathbf { x } \| _ { 2 } }$ ensures ${{\left\| \mu _{c}^{l} \right\|}_{2}}=1$. All class centroids are permanently stored after traning.

For a test example, extract its feature vectors at each layer $h _ { \text{test} } ^ { l }$ and compute the cosine similarities with all class centroids
\begin{equation} \label{eq:2}
s_{c}^{l} = \left\langle h_{{c,}\text{test}}^{l},\mu _{c}^{l} \right\rangle = \frac{h_{{c,}\text{test}}^{l} \cdot \mu_{c}^{l}}{\lVert h_{{c,}\text{test}}^{l} \rVert \lVert \mu_{c}^{l} \rVert} = h_{{c,}\text{test}}^{l} \cdot \mu _{c}^{l}.
\end{equation}
For the constructed layer-wise feature paths in Eq. \ref{eq:1}, we propose a hierarchical similarity fusion strategy that performs weighted aggregation of similarities across $L$ intermediate layers:
\begin{equation} \label{eq:3}
S _ { c } = \sum _ { l = 1 } ^ { L } \omega _ { l } s _ { c } ^ { l },
\end{equation}
where $\omega _ { l }$ denotes the weight parameter for layer $l$, and $S _ { c }$ represents the aggregated similarity score between the example and class $c$.

\subsection{Detection}
\label{section:22}
First for the adversarial example detection problem, we generate adversarial examples using AutoProjected Gradient Descent (AutoPGD) attacks \cite{14}. Given an original example $x$ and the target DNN model $f$, the adversarial example generation formula is as follows:
\begin{equation} \label{eq:4}
\begin{aligned}
& x_ { \mathrm { a d v } } = \arg \max _ { x ^ { \prime } } L \left( f \left( x ^ { \prime } \right) , y \right) \\
& \text{s.t.} \left\| x ^ { \prime } - x \right\| _ { p } \leq \varepsilon, \quad p \in \{ 0 , 1 , 2 , \infty \},
\end{aligned}
\end{equation}
where $L$ denotes the cross-entropy loss, and $\varepsilon$ is the maximum perturbation bound. Pixel-space projection and input normalization are applied during attacks to ensure adversarial examples satisfy the physical constraints of $\chi _ { \mathrm { a d v } } \in [ 0 , 1 ] ^ { D }$. $p$ denotes the attack norm, including $L_ { 0 }$, $L_ { 1 }$, $L_ { 2 }$, and $L_ { \infty \ }$ norms.

In order to minimize the feature collapse in the final few layers that leads to too high similarity, we remove the layers where PFC occurs to ensure the comparability of similarity. Considering that clean examples will definitely maintain a high similarity to its correct class, while adversarial examples will have a lower similarity to any class-centered path, the maximum similarity among all classes is valid as a distinguishing statistic. We then construct the detection statistic based on the aggregated similarity in Eq. \ref{eq:3}.
\begin{equation} \label{eq:5}
S _ { \max } = \max _ { c } S _ { c }.
\end{equation}
For each test example, compute its maximum similarity to all classes and compare it with a preset threshold $\tau$. If $S _ { \max }$ < $\tau$, classify $x$ as adversarial; otherwise, classify it as clean. We adopt the core principle of Otsu's thresholding method \cite{15}, determining $\tau$ by maximizing the variance between layer-wise feature paths of clean and adversarial examples.

\subsection{Recognition}
\label{section:23}
Based on the aforementioned detection results, we design an image recognition method for both clean and adversarial examples. Examples determined to be clean using the detection method are still directly recognized using the trained model. Meanwhile, adversarial examples are also recognized using layer-wise feature paths. Since adversarial examples features gradually change from their original correct class to the wrong class across network layers, the final few layers cannot be selected for classification. Therefore, we should rely on middle-layer decisions when they are correct, and avoid using the weakly robust final layers for reliable recognition.

To identify more robust intermediate layers, we statistically evaluate the adversarial classification accuracy across all layers and select the top-$L ^ { \prime }$ layers with higher accuracy. We implement a weighted voting mechanism for these $L ^ { \prime }$ intermediate layers to aggregate their predictions. To demonstrate the broad applicability of our approach, we intentionally avoid complex weighting strategies and assign equal weights to all selected intermediate layer outputs.
\begin{equation} \label{eq:6}
\operatorname { S c o r e } ( \mathrm { k } ) = \sum _ { l = 1 } ^ { L ^ { \prime } } \omega _ { l } \cdot \delta \left( \underset { m } { \arg \max }\, s _ { m } ^ { l } = k \right),
\end{equation}
where $\delta ( \cdot )$ denotes an indicator function that equals 1 if the condition holds and 0 otherwise. The predicted result $\hat { y }$ is the highest scoring result after weighted voting. If multiple layers predict the same result, select the prediction from the last layer in $L ^ { \prime }$.
\begin{equation} \label{eq:7}
\hat { y } = \arg \max _ { k } \operatorname { S c o r e } ( k ).
\end{equation}
\section{Experiment}
\label{section:3}
In this section, we present the ResNet-20 with PFC and the standard ResNet-18 on CIFAR-10 dataset, with their architectures detailed in Table \ref{tab:1}. We then perform adversarial example detection and image recognition on ResNet-20, demonstrating the superiority of the layer-wise feature paths in both tasks. Additionally, the study is extended to the standard ResNet-18 in Section \ref{section:32}.

\begin{table}[ht]
\centering
\caption{\text{Architecture specifications of ResNet-20 and ResNet-18}}
\label{tab:1}
\begin{tabular}{cccc}
\toprule
Model & K & Intermediate dim & blocks \\
\midrule
\multirow{3}{*}{ResNet-20} & \multirow{3}{*}{3} & 64$\times$16$\times$16 & 2 \\
 &  & 128$\times$8$\times$8 & 3 \\
 &  & 256$\times$4$\times$4 & 4 \\
\midrule
\multirow{4}{*}{ResNet-18} & \multirow{4}{*}{4} & 64$\times$16$\times$16 & 2 \\
 &  & 128$\times$8$\times$8 & 2 \\
 &  & 256$\times$4$\times$4 & 2 \\
 &  & 512$\times$2$\times$2 & 2 \\
\bottomrule
\end{tabular}
\end{table}

The experiments adopt the CIFAR-10 benchmark dataset. The size of the training set is 50,000 and the size of the test set is 10,000, which are divided into 10 classes. The training set is trained by the model to form layer-wise feature paths in the center of each class, while the test set is used to generate adversarial examples, then the test set which we use has 10,000 clean examples and 10,000 adversarial examples. The SGD optimizer is configured with a momentum of 0.9, weight decay of 5e-4, and batch size of 128. Training runs for 300 epochs, with an initial learning rate of 0.01 reduced by a factor of 10 at epochs 117 and 233. The AutoPGD attack uses a perturbation magnitude of 8/255, step size of 2/255, and 20 iterations. All experiments are conducted on an NVIDIA TITAN RTX GPU (24GB memory).

\subsection{Adversarial robustness of the network with PFC}
\label{section:31}
Building on the analysis in Section \ref{section:21}, the network with PFC exhibits more compact feature class centers and clearer layer-wise feature paths. To validate this hypothesis, we first investigate the adversarial robustness of ResNet-20.

\subsubsection{Adversarial example detection}

First, we compute the cosine similarity between the layer-wise feature paths of test examples and the layer-wise class center paths of the trained model. The average cosine similarity is calculated per class. Visualizations of the layer-wise cosine similarity paths reveal that clean examples exhibit high cosine similarity with their correct class centers, gradually increasing across layers in Fig. \ref{fig:311_1}. In contrast, adversarial examples show distinct deviations in feature paths, particularly after the emergence of the PFC phenomenon, which amplifies their divergence from the correct paths.

\begin{figure}[htbp]
\centering
\includegraphics[width=0.7\textwidth]{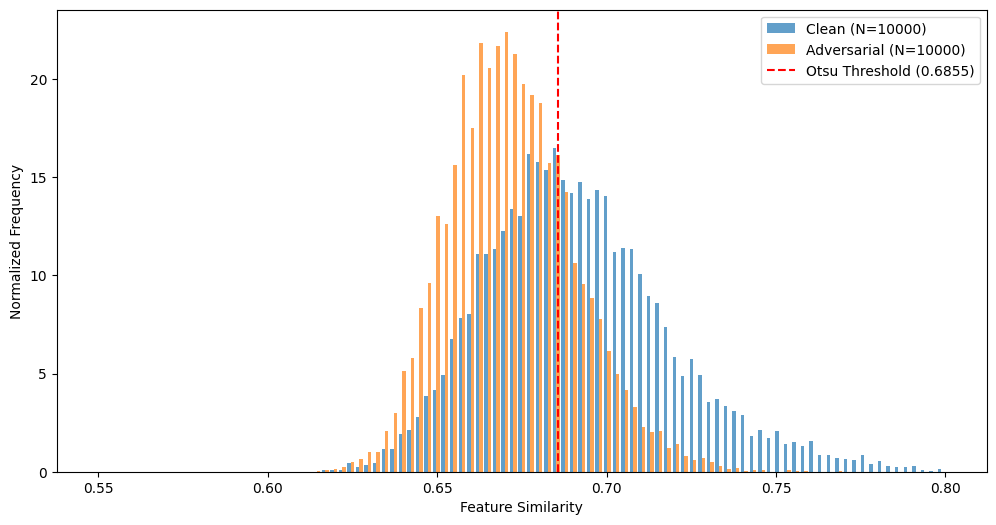}
\caption{Feature similarity distribution histogram of ResNet-20}
\label{fig:311_2}
\end{figure}

Clean and adversarial examples exhibit distinct differences and separability in their layer-wise feature paths, enabling threshold-based separation. The optimal threshold is determined by maximizing the variance of cosine similarities between clean and adversarial examples. As shown in Fig. \ref{fig:311_2}, the histogram illustrates the feature similarity distribution. At the optimal threshold $\tau=0.6855$, the rate that adversarial examples is successfully detected is 80.28\%.

The experimental results show that the layer-wise feature paths can effectively detect both clean and adversarial examples using only the network output features without any prior information and training strategy.

\subsubsection{Image recognition}

In order to verify the generalization of our method on the two tasks, we use the post-detection results as the dataset for the recognition task, i.e., detecting less than the threshold is determined as adversarial examples, and vice versa as clean examples. For the detected clean examples, the trained model is still employed to perform classification judgment. As for detected adversarial examples, Fig.\ref{fig:311_1} shows that they inevitably deviate from the correct class as network depth increases. And PFC exacerbates the degree of path away, i.e., after the phenomenon of PFC occurs in the middle layers, all subsequent layers rapidly move away from the correct class. Therefore, we cannot rely on the judgment of the final few layers and need to find intermediate layers which are robust.

\begin{figure}[htbp]
\centering
\includegraphics[width=0.7\textwidth]{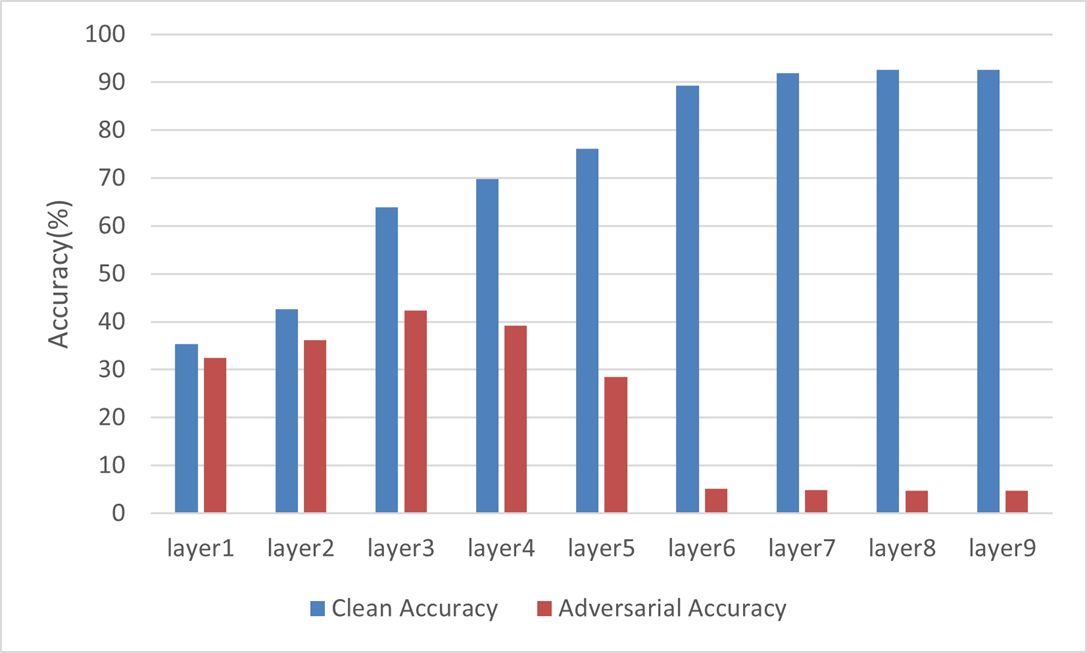}
\caption{Layer-wise feature classification accuracy of ResNet-20 on the initial examples}
\label{fig:312_1}
\end{figure}

By analyzing layer-wise feature paths, we directly determine the nearest class as the result at each layer without classifier inference, significantly reducing computational costs. As shown in Fig. \ref{fig:312_1}, we count the results of classifying clean and adversarial examples using feature similarity at each layer on the initial examples. Layers 2, 3, and 4 achieve classification accuracies above 35\%. We then perform the recognition based on the detected examples and experiments demonstrate that a weighted combination of layers 2, 3, and 4 achieves the highest classification accuracy of 44.17\% in Table \ref{tab:3}.

To demonstrate that our method maintains high robustness without compromising clean accuracy, we compare it with standard training and adversarial training. As shown in Table \ref{tab:31}, our method achieves 82.77\% clean accuracy and 44.17\% adversarial accuracy. Compared to standard training, adversarial accuracy increased from 5\% to 44\%, changing the previous perception that neural networks are weakly robust. While maintaining comparable performance to adversarial training with 77.64\% clean accuracy and 52.94\% adversarial accuracy, our method is an acceptable trade-off and requires no additional computational overhead.

\begin{table}[ht]
\centering
\caption{Image recognition accuracy on ResNet-20}
\begin{tabular}{@{}c cc@{}}
\toprule
\multirow{2}{*}{\centering Method} & \multicolumn{2}{c}{ResNet-20} \\ 
\cmidrule(lr){2-3}
& Clean Accuracy & Adversarial Accuracy \\
\midrule
Standard Training & 92.03 & 4.91 \\
Adversarial Training & 77.64 & 52.94 \\
Layer-wise Feature Paths & 82.77 & 44.17 \\
\bottomrule
\end{tabular}
\label{tab:31}
\end{table}

\subsection{Adversarial robustness of the standard ResNet-18}
\label{section:32}
To validate the generalizability of our method, we conduct additional experiments on the standard ResNet-18, following our investigation of the characteristics of ResNet-20 with PFC. The standard ResNet-18 has not been specially parameterized and does not have significant feature clustering.

\subsubsection{Adversarial example detection}
\label{section:321}
\begin{figure}[htbp]
\centering
\includegraphics[width=0.7\textwidth]{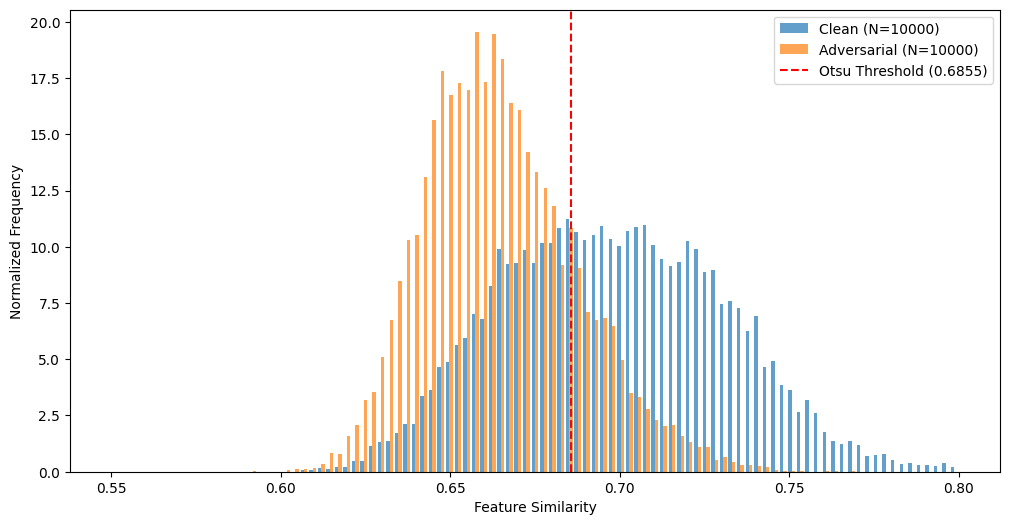}
\caption{Feature similarity distribution histogram of ResNet-18}
\label{fig:321_2}
\end{figure}

We used the same method and parameter settings for adversarial example detection on ResNet-18. Although the PFC network demonstrated tighter clustering of feature class centers, ResNet-18 still achieved equivalent detection performance with 82.8\% accuracy through optimal threshold selection in Fig. \ref{fig:321_2}.


Interestingly, ResNet-18 has the same detection threshold $\tau=0.6855$ as ResNet-20, suggesting that small differences in network structure do not have a significant effect on the similarity between layer-wise feature paths.

The standard ResNet-18 still achieves good results on the detection task, indicating that the layer-wise feature paths method does not depend on well clustered networks and remains generalizable on normal networks.

\subsubsection{Image recognition}

\begin{figure}[htbp]
\centering
\includegraphics[width=0.7\textwidth]{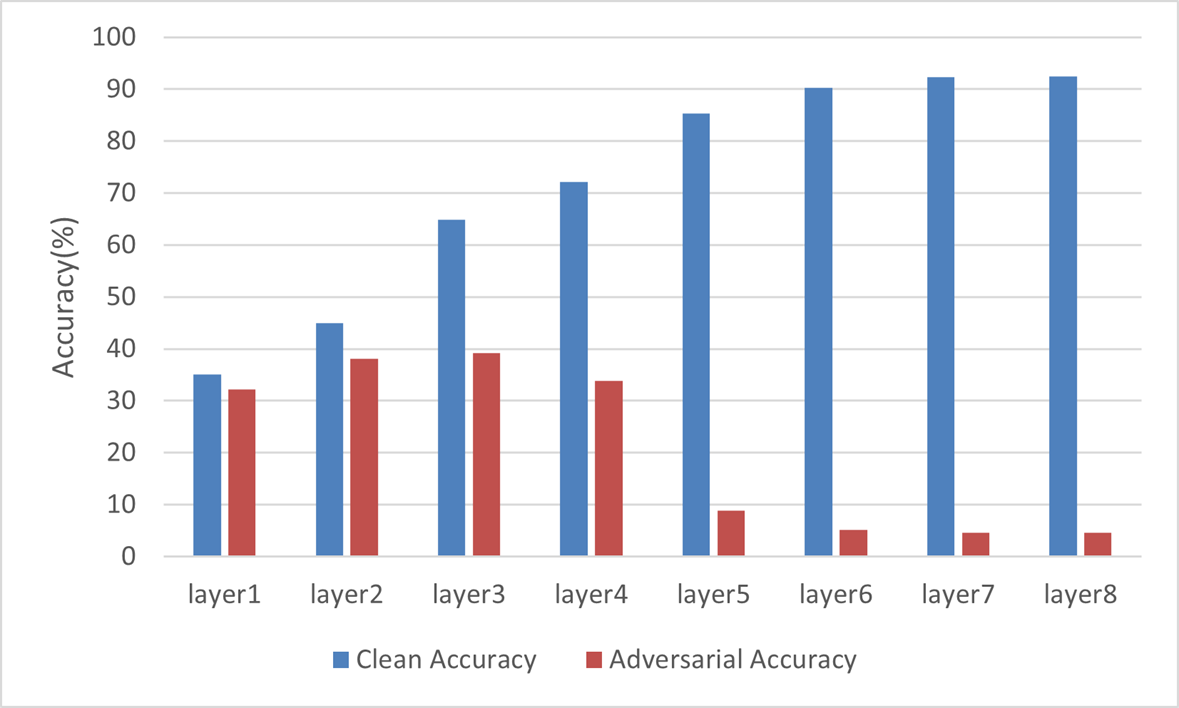}
\caption{Layer-wise feature classification accuracy of ResNet-18 on the initial examples}
\label{fig:322_1}
\end{figure}

Section \ref{section:321} confirms the adversarial example detection capability of ResNet-18. Similarly, we have conducted experiments for image recognition, in which the middle layers still have high robustness, as shown in Fig. \ref{fig:322_1}. To illustrate the generalizability of our method, the same middle layers are taken for classification and the adversarial accuracy can reach 46.1\% in Table \ref{tab:3}. This result also far exceeds the 4.9\% adversarial accuracy under standard training of ResNet-18, achieving a trade-off without adding additional computational overhead  in Table \ref{tab:322}.

\begin{table}[H]
\caption{\text{Recognition results for different middle layers configurations of two models}}
\centering
\begin{tabular}{ccccc}
\toprule
Layer&layer2+3&layer2+4&layer3+4&layer2+3+4 \\
\midrule
ResNet-20&37.16&36.26&43&44.17 \\
ResNet-18&37.94&38.47&45.12 &46.1  \\
\bottomrule
\end{tabular}
\label{tab:3}
\end{table}

Extending our method to standard ResNet-18 achieves comparable performance to ResNet-20 with PFC in both adversarial example detection and image recognition, while significantly surpassing standard training in adversarial robustness. The standard ResNet-18 has good performance on the recognition task, again verifying that PFC is not a critical influence. Compared to adversarial training, the method in this paper does not require any prior information and high computational overhead. Compared to standard training, experimental results on ResNet-18 demonstrate that classical neural networks are robust.

\begin{table}[ht]
\centering
\caption{Image recognition accuracy on ResNet-18}
\begin{tabular}{ccc}
\toprule
\multirow{2}{*}{Method} & \multicolumn{2}{c}{ResNet-18} \\ 
\cmidrule(lr){2-3}
& Clean Accuracy & Adversarial Accuracy \\
\midrule
Standard Training & 92.24 & 4.9 \\
Adversarial Training & 78.48 & 50.74 \\
Layer-wise Feature Paths & 80.01 & 46.1 \\
\bottomrule
\end{tabular}
\label{tab:322}
\end{table}

\section{Conclusion}
In this work, we introduce a method for adversarial example detection and image recognition based on layer-wise feature paths of DNNs, exploiting the intrinsic adversarial robustness of DNN. Specifically, we compare the similarity of clean examples and adversarial examples to the layer-wise features of the class center. Experimental results demonstrate the two tasks achieve good performance on the network with PFC. Furthermore, extension experiments on standard ResNet-18 verify the inherent robustness embedded in DNN structures, with clean and adversarial example recognition accuracies reaching 80.01\% and 46.1\% respectively. Compared to adversarial training with 78.48\% clean accuracy and 50.74\% adversarial accuracy, our method achieves adversarial robustness similar to it, while spending less computational overhead. Our method is a new understanding of the robustness of deep neural networks and challenges the previous perception that deep neural networks are weakly robust.

\clearpage
\bibliographystyle{unsrt}
\bibliography{reference}

\begin{thebibliography}{10}

\bibitem{1}
Christian Szegedy, Wojciech Zaremba, Ilya Sutskever, Joan Bruna, Dumitru Erhan, Ian Goodfellow, and Rob Fergus.
\newblock Intriguing properties of neural networks.
\newblock {\em Computer Science}, 2013.

\bibitem{2}
Seyed-Mohsen Moosavi-Dezfooli, Alhussein Fawzi, Omar Fawzi, and Pascal Frossard.
\newblock Universal adversarial perturbations.
\newblock In {\em Proceedings of the IEEE conference on computer vision and pattern recognition}, pages 1765--1773, 2017.

\bibitem{3}
Ian~J. Goodfellow, Jonathon Shlens, and Christian Szegedy.
\newblock Explaining and harnessing adversarial examples.
\newblock {\em Computer Science}, 2014.

\bibitem{4}
Florian Tram{\`e}r, Alexey Kurakin, Nicolas Papernot, Ian Goodfellow, Dan Boneh, and Patrick McDaniel.
\newblock Ensemble adversarial training: Attacks and defenses.
\newblock {\em arXiv preprint arXiv:1705.07204}, 2017.

\bibitem{5}
Aleksander Madry, Aleksandar Makelov, Ludwig Schmidt, Dimitris Tsipras, and Adrian Vladu.
\newblock Towards deep learning models resistant to adversarial attacks.
\newblock {\em arXiv preprint arXiv:1706.06083}, 2017.

\bibitem{6}
Nicholas Carlini and David Wagner.
\newblock Towards evaluating the robustness of neural networks.
\newblock In {\em 2017 ieee symposium on security and privacy (sp)}, pages 39--57. Ieee, 2017.

\bibitem{7}
Nicolas Papernot, Patrick McDaniel, Xi~Wu, Somesh Jha, and Ananthram Swami.
\newblock Distillation as a defense to adversarial perturbations against deep neural networks.
\newblock In {\em 2016 IEEE symposium on security and privacy (SP)}, pages 582--597. IEEE, 2016.

\bibitem{8}
Seyed-Mohsen Moosavi-Dezfooli, Alhussein Fawzi, and Pascal Frossard.
\newblock Deepfool: a simple and accurate method to fool deep neural networks.
\newblock In {\em Proceedings of the IEEE conference on computer vision and pattern recognition}, pages 2574--2582, 2016.

\bibitem{9}
Andrew Ilyas, Logan Engstrom, Shibani Santurkar, Brandon Tran, Dimitris Tsipras, and Aleksander Madry.
\newblock Adversarial examples are not bugs, they are features.
\newblock In {\em Advances in Neural Information Processing Systems 32, Volume 1 of 20: 32nd Conference on Neural Information Processing Systems (NeurIPS 2019).Vancouver(CA).8-14 December 2019}, 2020.

\bibitem{10}
Skander Karkar, Patrick Gallinari, and Alain Rakotomamonjy.
\newblock Adversarial sample detection through neural network transport dynamics.
\newblock In {\em Joint European Conference on Machine Learning and Knowledge Discovery in Databases}, pages 164--181. Springer, 2023.

\bibitem{11}
Sicong Wang, Kuo Gai, and Shihua Zhang.
\newblock Progressive feedforward collapse of resnet training.
\newblock {\em arXiv preprint arXiv:2405.00985}, 2024.

\bibitem{12}
Vardan Papyan, X~Y Han, and David~L Donoho.
\newblock Prevalence of neural collapse during the terminal phase of deep learning training.
\newblock {\em Proceedings of the National Academy of Sciences}, 2020.

\bibitem{13}
Hangfeng He and Weijie~J Su.
\newblock A law of data separation in deep learning.
\newblock {\em Proceedings of the National Academy of Sciences}, 120(36):e2221704120, 2023.

\bibitem{14}
Francesco Croce and Matthias Hein.
\newblock Reliable evaluation of adversarial robustness with an ensemble of diverse parameter-free attacks.
\newblock In {\em 37th International Conference on Machine Learning: ICML 2020, Online, 13-18 July 2020, Part 3 of 15}, 2021.

\bibitem{15}
Nobuyuki Otsu.
\newblock A threshold selection method from gray-level histograms.
\newblock {\em IEEE Transactions on Systems Man \& Cybernetics}, 9(1):62--66, 2007.

\end{thebibliography}

\appendix

\end{document}